\documentclass[11pt]{article}

\usepackage[final]{acl}

\usepackage{times}
\usepackage{latexsym}
\usepackage[T1]{fontenc}
\usepackage{hyperref}
\usepackage{url}
\usepackage{natbib}
\usepackage{amsmath, amssymb}
\usepackage{booktabs}
\usepackage[utf8]{inputenc}
\usepackage{graphicx}
\usepackage{xcolor}

\newcommand{\AMDGPU}{AMD~Instinct\textsuperscript{\texttrademark}}
\usepackage{microtype}

\usepackage{inconsolata}

\usepackage{graphicx}

\title{AdaptEvolve: Improving Efficiency of Evolutionary AI Agents through Adaptive Model Selection}

\author{\textbf{Pretam Ray}\textsuperscript{$\dagger$}\thanks{Work done during Internship at Advanced Micro Devices, Inc. (AMD)},
\textbf{ Pratik Prabhanjan Brahma}\textsuperscript{$\diamond$},\textbf{ Zicheng Liu}\textsuperscript{$\diamond$} \textbf{and} \textbf{ Emad Barsoum}\textsuperscript{$\diamond$}
  \\
  \textsuperscript{$\dagger$} IIT Kharagpur,
  \textsuperscript{$\diamond$} Advanced Micro Devices, Inc. (AMD) \\
  }

\begin{document}
\maketitle
\begin{abstract}
Evolutionary agentic systems intensify the trade-off between computational efficiency and reasoning capability by repeatedly invoking large language models (LLMs) during inference.This setting raises a central question: \emph{how can an agent dynamically select an LLM that is sufficiently capable for the current generation step while remaining computationally efficient?} While model cascades offer a practical mechanism for balancing this trade-off, existing routing strategies typically rely on static heuristics or external controllers and do not explicitly account for model uncertainty. We introduce \emph{ AdaptEvolve: Adaptive LLM Selection for Multi-LLM Evolutionary Refinement} within an evolutionary sequential refinement framework that leverages intrinsic generation confidence to estimate real-time solvability. Empirical results show that confidence-driven selection yields a favorable Pareto frontier, reducing total inference cost by an average of \textbf{37.9\%} across benchmarks while retaining \textbf{97.5\%} of the upper-bound accuracy of static large-model baselines. Our codebase is available at \href{https://github.com/raypretam/adaptive_llm_selection}{AdaptEvolve}.

\end{abstract}

\section{Introduction}
\begin{figure*}[t] 
    \centering
    \includegraphics[width=0.70\textwidth]{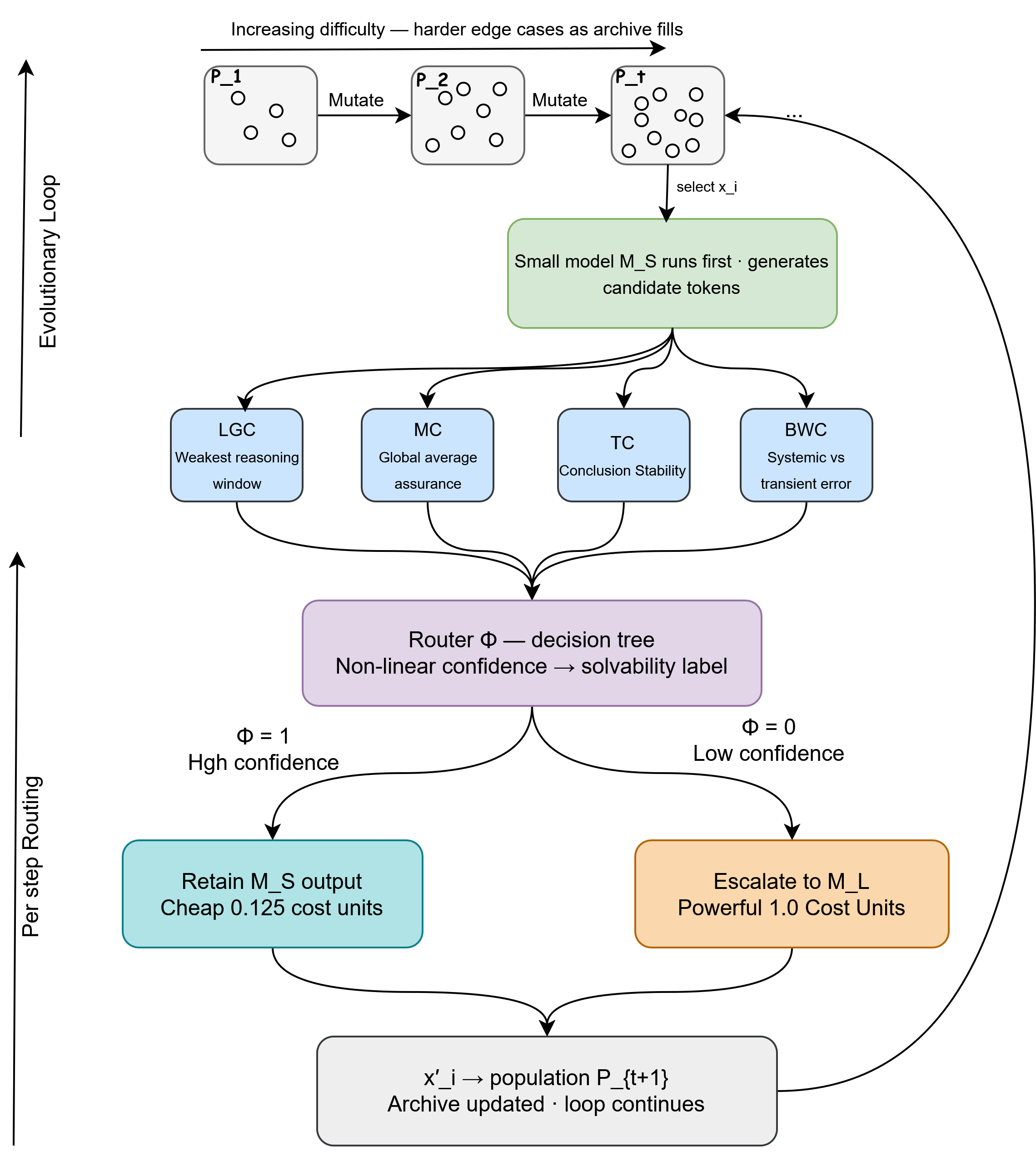}
    \caption{\textbf{Adaptive Evolutionary Refinement Framework.} The workflow initiates with a candidate generation using the small model ($\mathcal{M}_S$). We compute intrinsic confidence metrics (LGC, MC, TC, BWC) based on token entropy. A lightweight decision tree router ($\Phi$), bootstrapped via a warm-up phase, dynamically determines whether to retain the efficient generation or escalate to the large model ($\mathcal{M}_L$) for complex reasoning hurdles.}
    \label{fig:framework}
\end{figure*}

The computational cost of state-of-the-art large language models (LLMs) remains a key barrier to scalable deployment. While recent advances in small-model distillation \cite{grattafiori2024llama, abdin2024phi} and self-consistency mechanisms \cite{wang2022self} have substantially improved the capabilities of efficient models in the 1–4B parameter range, a persistent performance gap relative to frontier models continues to limit their standalone applicability in complex reasoning tasks.

At the same time, agentic reasoning paradigms have evolved from single-turn generation toward \emph{iterative evolutionary refinement} involving multiple LLMs, as exemplified by AlphaEvolve \cite{novikov2025alphaevolvecodingagentscientific}, OpenEvolve \cite{openevolve}, and ShinkaEvolve \cite{lange2025shinka}. These frameworks iteratively generate, mutate, and select candidate solutions using multiple models. Also, GEAK-OptimAgent\_V2 \cite{wang2025geakintroducingtritonkernel} uses OpenEvolve framework for generating Triton kernels. However, model usage within such systems is typically governed by fixed sampling weights or predetermined schedules, without explicit consideration of which model is most appropriate for a given refinement step. Recent efforts have explored accelerating agentic systems through improved planning and search strategies \cite{yu2025alpharesearchacceleratingnewalgorithm, wu2025gapgraphbasedagentplanning, lange2025shinka}, while parallel lines of work have studied \emph{multi-model inference systems} that explicitly optimize the quality-cost trade-off predominantly fall into three paradigms: \emph{query routing}, where a classifier predicts input difficulty and dispatches requests accordingly \cite{ong2024routellmlearningroutellms}; \emph{speculative decoding}, which employs smaller models to draft candidate tokens for verification by larger models \cite{kim2023speculative}; and \emph{model cascades}, which progressively escalate queries from lower-cost to higher-capacity models \cite{chen2023frugalgpt}. Concurrent to our work, \citet{ding2025bestrouteadaptivellmrouting} propose BEST-Route, which also  explores test-time adaptive routing for LLMs; our work is distinguished by its application to the non-stationary, population-level difficulty distributions inherent in evolutionary agentic refinement.

In this work, we draw inspiration from multi-model inference systems to accelerate evolutionary agentic coding frameworks composed of multiple LLMs. We argue that effective model selection in this setting should be driven by \emph{intrinsic uncertainty signals} rather than external routing models. Prior work has shown that entropy and logit-based measures expose systematic confidence signatures that correlate with model failure modes and reasoning breakdowns \cite{fu2025deepthinkconfidence, liu-etal-2024-llms-learn-uncertainty}. These signals are readily available at inference time and incur negligible additional computational cost.

Building on these observations, we propose \emph{AdaptEvolve}, a framework that conditions model allocation on intrinsic generation uncertainty within evolutionary agentic settings. This creates a dynamic decision process where efficient models handle routine tasks, and powerful models are selectively invoked for high-entropy steps. To our knowledge, this is the first application of uncertainty-aware adaptive selection in evolutionary agentic refinement.

Our contributions are:
\begin{itemize}
    \setlength\itemsep{0em}
    \item \textbf{Lightweight Calibration:} We introduce a resource-efficient calibration method that bootstraps a shallow decision tree using intrinsic entropy metrics \cite{fu2025deepthinkconfidence} from a minimal warm-up set ($N=50$), eliminating the need for heavyweight routers.
    \item \textbf{Pareto Efficiency:} Empirical evaluation demonstrates a superior Pareto frontier, reducing total inference compute by \textbf{37.9\%} while retaining \textbf{97.5\%} of the upper-bound performance of static large-model baselines.
\end{itemize}

\section{Methodology}
\begin{table*}[t]
\centering
\resizebox{1.0\textwidth}{!}{%
\begin{tabular}{@{}l cccc cccc@{}}
\toprule
& \multicolumn{4}{c}{\textbf{LiveCodeBench V5}} & \multicolumn{4}{c}{\textbf{MBPP}} \\
\cmidrule(lr){2-5} \cmidrule(lr){6-9}
\textbf{Configuration} & \textbf{Ratio (S:L)} & \textbf{Cost} & \textbf{Acc} & \textbf{Eff.} & \textbf{Ratio (S:L)} & \textbf{Cost} & \textbf{Acc} & \textbf{Eff.} \\ 
\midrule

\multicolumn{9}{l}{\textit{\textbf{Baselines}}} \\
4B Iterative (Lower Bound) & 100:0 & 0.46 & 62.3 & 135.4 & 100:0 & 0.37 & 80.1 & 216.5 \\
32B Iterative (Upper Bound) & 0:100 & 3.17 & 75.2 & 23.7 & 0:100 & 1.18 & 94.0 & 79.7 \\ 
\midrule

\multicolumn{9}{l}{\textit{\textbf{Static Routing}}} \\
\textbf{Decision Tree (Gini, Depth=5)} & 43:57 & 2.02 & 71.2 & 35.2 & 96:04 & 0.46 & 82.2 & 178.7 \\
Random Sampling & 43:57 & 1.98 & 67.5 & 34.1 & 96:04 & 0.45 & 81.3 & 180.7 \\ 
\midrule

\multicolumn{9}{l}{\textit{\textbf{Online Adaptation}}} \\
\textbf{Adaptive Hoeffding Tree \cite{adapt_tree}} & \textbf{42:58} & \textbf{2.08} & \textbf{73.6} & \textbf{35.4} & \textbf{85:15} & \textbf{0.69} & \textbf{91.3} & \textbf{132.3} \\
Random Sampling (Online Ratio) & 42:58 & 2.01 & 68.2 & 33.9 & 85:15 & 0.68 & 88.3 & 129.9 \\ 
\bottomrule
\end{tabular}%
}
\caption{\textbf{Comparative Analysis of Adaptive Routing across Benchmarks.} We evaluate cost-efficiency trade-offs. \textit{Cost} denotes Compute Cost (one 32B call = 1 units and one 4B call = 0.125 units). \textit{Eff} (\textbf{Efficiency} ($\text{Eff} = \text{Accuracy} / \text{Compute Cost}$)).}

\label{tab:results_combined}
\end{table*}

\subsection{Adaptive Evolutionary Refinement}
Our framework builds upon OpenEvolve \cite{openevolve}, which iteratively optimizes solution candidates, represented as code generations $P_t$, through mutation and selection. Standard evolutionary approaches utilize a fixed model for all mutations $\mathcal{M}$, ignoring the dynamic variance in reasoning difficulty. We propose \textbf{AdaptEvolve: Adaptive Evolutionary Refinement}, transforming the mutation operator into a conditional routing function.

Let $\mathcal{M}_S$ and $\mathcal{M}_L$ denote a cost-efficient model (e.g., 4B) and a capability-dense model (e.g., 32B), respectively. For a candidate $x_i \in P_t$, we compute a confidence vector $C(x_i)$ from the generation statistics of $\mathcal{M}_S$. The mutation for the subsequent generation is defined as:
\begin{equation}
    x_{i}' = \begin{cases} 
    \mathcal{M}_S(x_i) & \text{if } \Phi(C(x_i)) = 1 \\
    \mathcal{M}_L(x_i) & \text{otherwise}
    \end{cases}
\end{equation}
where $\Phi: \mathbb{R}^d \to \{0,1\}$ is a lightweight binary classifier predicting solvability. This mechanism creates a dynamic computation graph where $\mathcal{M}_S$ handles routine refinement steps, and $\mathcal{M}_L$ is reserved for high-entropy reasoning hurdles.

A key property distinguishing evolutionary search from generic multi-step generation is \emph{population-level concept drift}: as the MAP-Elites archive fills with high-scoring solutions, subsequent mutations must address increasingly difficult edge cases, causing the difficulty distribution to shift non-stationary over time. Static routers degrade in this setting, as evidenced by the 2.4-point accuracy gap between the static Decision Tree and HAT on LiveCodeBench (Table~\ref{tab:results_combined}). To our knowledge, AdaptEvolve is the first routing framework to address this  non-stationarity via test-time online adaptation, distinguishing it from prior single-step routing approaches such as RouteLLM~\citep{ong_routellm_2025}.




\subsection{Entropy-Based Confidence Metrics}
\label{sec:conf_metrics}

We quantify generation uncertainty via \textit{Token Confidence} ($c_t$), defined as the negative mean log-probability of the top-$k$ tokens at position i \cite{fu2025deepthinkconfidence}:
\begin{equation}
    c_i = - \frac{1}{k} \sum_{j=1}^{k} \log P_i(j)
\end{equation}
High confidence corresponds to peaked distributions and greater model certainty, while low confidence indicates uncertainty in token prediction. We also use four scalar metrics \cite{fu2025deepthinkconfidence} over the sequence for uncertainty quantification:

\begin{itemize}
    \setlength\itemsep{0em}
    \item \textbf{Mean Confidence (MC):} The global average uncertainty ($\frac{1}{T} \sum c_i$), serving as a proxy for overall model assurance.
    \item \textbf{Lowest Group Confidence (LGC):} The window with the maximum uncertainty score ($\min G_i$), identifying the localized "weakest link" in the reasoning chain.
    \item \textbf{Tail Confidence (TC):} The uncertainty of the final $W$ tokens ($G_{T-W+1}$), targeting the stability of the solution's conclusion.
    \item \textbf{Bottom-K\% Confidence (BWC):} The average of the highest $K$-th percentile of window scores, distinguishing systemic hallucination from transient token noise.
\end{itemize}
\section{Experiment}
\subsection{Experimental Setup}
We implement our adaptive framework on top of OpenEvolve \cite{openevolve}, an open-source evolutionary framework inspired by AlphaEvolve \cite{novikov2025alphaevolvecodingagentscientific}, an AI-coding Agent, which performs iterative population-based refinement of candidate solutions using large language models as mutation and evaluation operators. The experiments were executed on a single node with 8× \AMDGPU MI250 GPUs. The code generations are dynamically updated using a map-elites \cite{mouret2015illuminatingsearchspacesmapping} style archive, where previous high-scoring generations serve as few-shot exemplars for the next step. We define hyperparameters of the system in \autoref{sec:hyperparameters}.

\subsection{Models}
We evaluate the following multi-LLM system from the Qwen3 family \cite{qwen3} known for strong reasoning capabilities: Qwen3-4B (Small/Router Target) and Qwen3-32B (Large/Escalation Target).

\subsection{Dataset and Metrics}
We select coding benchmarks that provide deterministic ground-truth signals also due to the fact that Openevolve generates code as solution representations.
\begin{itemize}
    \item LiveCodeBench v5 \cite{jain2024livecodebench}: A challenging code generation benchmark requiring syntax correctness and functional logic, consisting of 880 samples.
    \item MBPP \cite{austin2021program}: The benchmark consists of around 1,000 crowd-sourced Python programming problems, consisting of 974 samples.
\end{itemize}

\paragraph{Metrics:} We evaluate performance using accuracy, assigning a binary score of 1 to generated solutions that pass all test cases and 0 otherwise.

\paragraph{Cost Metric Validity.}
We normalize inference cost as: $\text{Cost} \propto \text{Parameters} \times \text{Tokens}$. Since our setup enforces a strict 20{,}000-token maximum per call (Table~\ref{tab:config_params_updated}) and empirically yields matched mean generation lengths ($\sim$15K tokens) across all configurations with no systematic token inflation from smaller models, total token counts 
are effectively controlled. Under these constraints, model size becomes the dominant scaling factor, making the call-count normalization (32B\,$=$\,1, 4B\,$=$\,0.125) a faithful proxy for actual FLOPs. Wall-clock measurements on LLaMA~3.1 / HumanEval confirm close alignment between normalized cost and real inference time (§\ref{sec:results}).

\begin{table}[h]
\centering
\resizebox{\columnwidth}{!}{%
\begin{tabular}{@{}lcccc@{}}
\toprule
\textbf{Configuration} & \textbf{Ratio (S:L)} & \textbf{Cost (Norm.)} & \textbf{Accuracy} & \textbf{Eff.} \\ \midrule
Cascading Baseline \cite{chen2023frugalgpt} & 36:64 & 2.81 & 73.8 & 26.3 \\
\textbf{Adaptive Hoeffding Tree (Ours)} & \textbf{42:58} & \textbf{2.08} & \textbf{73.6} & \textbf{35.4} \\ \bottomrule
\end{tabular}%
}
\caption{\textbf{Comparison with Cascading.} The Adaptive Hoeffding Tree achieves a superior Pareto trade-off. While the raw accuracy is comparable, our method demonstrates significantly higher Efficiency (\textbf{35.4} vs 26.3) due to intelligent routing that lowers compute cost by 0.73.}
\label{tab:cascading_comparison}
\end{table}
\subsection{Baselines}
We compare our Adaptive Confidence Router against both static and heuristic baselines:
\begin{itemize}
    \item \textbf{Pure Small / Pure Large:} The lower and upper bounds of performance and cost.
    \item \textbf{Random Routing:} Randomly selecting models to establish a chance baseline.
    \item \textbf{Cascade:} We start with the smaller model always and as soon as the smaller model fails we defer to the bigger model.
\end{itemize}

\subsection{Methodology}
\label{sec:methodology}

The mapping from confidence signatures to solvability is non-linear and non-stationary. A simple threshold on Mean Confidence often fails to detect confident hallucinations (low MC, high LGC) \autoref{sec:stat_analysis}

\paragraph{Static Decision Tree}
To bootstrap $\Phi$, we execute a minimal warm-up phase ($N = 50$) using $\mathcal{M}_S$ and $\mathcal{M}_L$ to generate a labeled dataset $\{(C(x_i), y_i)\}$, where $y_i$ denotes whether that sample will be solved by the model or not (Table~\ref{tab:router_logic}). Each warm-up instance corresponds to one candidate solution generated by $\mathcal{M}_S$ for a unique problem, paired with a ground-truth solvability label. The confidence vector $C(x_i)$ is computed over the full token sequence of that generation. Instances 
are drawn via stratified sampling that preserves the difficulty-level distribution of the source benchmark, ensuring representative coverage despite the small sample size. We train an initial DecisionTree~\citep{sklearn_api} (Gini impurity, max\_depth=5) on this subset, enabling inference that captures non-linear interactions between metrics (e.g., low MC but high LGC).

\paragraph{Online Adaptation}
To handle the non-stationary nature of evolutionary search, we employ a \textbf{Hoeffding Adaptive Tree (HAT)} \cite{adapt_tree} using \cite{montiel2021river} package.
HAT incrementally updates split criteria and monitors leaf error rates. Upon detecting concept drift, it automatically prunes and regrows decision branches, enabling the router to dynamically recalibrate escalation thresholds as reasoning difficulty evolves.
\subsection{Results and Analysis}
\label{sec:results}

We quantify agent acceleration using \textbf{Efficiency} ($\text{Eff} = \text{Accuracy} / \mathcal{C}_{\text{total}}$), where $\mathcal{C}_{\text{total}}$ represents the cumulative inference cost. We standardize computational units such that a single invocation of the 32B model incurs a cost of $1.0$, while the efficient 4B model incurs a cost of $0.125$. As shown in \autoref{tab:results_combined}, \textbf{AdaptEvolve} establishes a superior Pareto frontier across both difficulty regimes. \autoref{fig:accuracy_vs_32b}

\paragraph{Benchmark Performance}
On \textit{LiveCodeBench}, our method retains \textbf{97.9\%} of the 32B upper-bound accuracy (73.6\% vs 75.2\%) while reducing compute cost by \textbf{34.4\%}. This yields an efficiency score of \textbf{35.4}, significantly surpassing the static 32B model (23.7) and the Cascading baseline (26.3; see \autoref{tab:cascading_comparison}). The acceleration is magnified on \textit{MBPP}, where the router identifies that 85\% of queries are solvable by the small model. This reduces cost by \textbf{41.5\%} while maintaining \textbf{97.1\%} of peak accuracy, resulting in an efficiency score of \textbf{132.3}, nearly double that of the pure large model (79.7).

\paragraph{Generalization Across Model Families.}
To validate that AdaptEvolve is not restricted to the Qwen3 family, we evaluate on LLaMA~3.1 models~\citep{grattafiori2024llama} using the HumanEval benchmark~\citep{chen2021codex}. As shown in Table~\ref{tab:llama_results}, the Adaptive Hoeffding Tree consistently outperforms random sampling under identical routing ratios (38:62), achieving 74.3\% accuracy versus 69.4\% for random selection, while attaining a $1.56\times$ speedup over the 70B iterative baseline. These results confirm that the confidence-driven routing strategy generalizes across architectures without modification.

\begin{table}[h]
\centering
\small
\resizebox{\columnwidth}{!}{%
\begin{tabular}{lcccc}
\toprule
\textbf{Configuration} & \textbf{Ratio (S:L)} & \textbf{Cost} & \textbf{Acc.} & \textbf{Speedup} \\
\midrule
8B Iterative   & 100:0  & 0.54 & 64.3 & $3.33\times$ \\
70B Iterative  & 0:100  & 4.16 & 77.1 & $1\times$ \\
\midrule
HAT (Ours)         & 38:62  & 3.61 & 74.3 & $1.56\times$ \\
Random Sampling    & 38:62  & 3.57 & 69.4 & $1.15\times$ \\
\bottomrule
\end{tabular}
}
\caption{Generalization to LLaMA~3.1 models on HumanEval. HAT outperforms random sampling at the same routing ratio, confirming model-agnostic applicability of AdaptEvolve. Speedup is relative to the 70B iterative baseline.}
\label{tab:llama_results}
\end{table}


\paragraph{Routing Dynamics}
We analyze the impact of adaptation mechanisms:
\begin{itemize}
    \setlength\itemsep{0em}
    \item \textbf{Adaptation to Drift:} The Adaptive Hoeffding Tree \cite{adapt_tree} strictly outperforms static decision trees. On LiveCodeBench, it improves accuracy by 2.4 points (73.6\% vs 71.2\%) over the static baseline by dynamically adjusting decision boundaries as the population evolves toward complex edge cases.
    \item \textbf{Overall Efficiency:} Across benchmarks, uncertainty-aware routing reduces inference compute by an average of \textbf{37.9\%} while recovering \textbf{97.5\%} of the upper-bound performance, confirming that intrinsic confidence provides a robust signal for safe agentic acceleration.
\end{itemize}


\section{Conclusion}
We introduced \emph{AdaptEvolve}, a confidence-driven routing framework that accelerates evolutionary agentic coding systems by optimizing the cost-capability trade-off at each refinement step. By conditioning model selection on intrinsic generation uncertainty, our approach eliminates the need for heavy external controllers. Empirical evaluation demonstrates that this strategy is highly effective: it reduces total inference compute by \textbf{37.9\%} while retaining \textbf{97.5\%} of the upper-bound accuracy. These results confirm that intelligent resource allocation is a viable pathway for scalable agentic reasoning.

\section{Limitations}
Our evaluation is restricted to coding benchmarks (LiveCodeBench, MBPP, HumanEval), as AdaptEvolve builds upon OpenEvolve~\citep{openevolve}, which represents candidate solutions as executable code and uses test-case pass/fail signals as the evolutionary fitness metric. This architectural assumption means the framework cannot be directly applied to mathematical or open-ended reasoning benchmarks where smaller models generate direct symbolic answers rather than executable representations, breaking the evolutionary loop's fitness evaluation.

\bibliography{custom}

\appendix

\section{Appendix}
\label{sec:appendix}
\begin{table*}[ht]
\centering
\resizebox{\textwidth}{!}{%
\begin{tabular}{lcccccccc}
\toprule
 & \multicolumn{2}{c}{\textbf{LGC}} & \multicolumn{2}{c}{\textbf{MC}} & \multicolumn{2}{c}{\textbf{TC}} & \multicolumn{2}{c}{\textbf{BWC}} \\ 
\cmidrule(lr){2-3} \cmidrule(lr){4-5} \cmidrule(lr){6-7} \cmidrule(lr){8-9}
\textbf{Statistic} & \textbf{Unsolved (0)} & \textbf{Solved (1)} & \textbf{Unsolved (0)} & \textbf{Solved (1)} & \textbf{Unsolved (0)} & \textbf{Solved (1)} & \textbf{Unsolved (0)} & \textbf{Solved (1)} \\ \midrule
\textbf{Mean} & 5.858 & 7.982 & 7.649 & 9.761 & 8.509 & 10.285 & 6.373 & 8.732 \\
\textbf{Std Dev} & 1.535 & 2.570 & 1.756 & 1.981 & 1.839 & 1.858 & 1.791 & 2.522 \\
\textbf{Skew} & 2.044 & 0.935 & 0.790 & 0.698 & 0.533 & 0.205 & 1.708 & 0.720 \\ \bottomrule
\end{tabular}%
}
\caption{\textbf{Statistical Separability of Confidence Metrics.} A comparison of generation statistics (Mean, Standard Deviation, and Skewness) between Unsolved (0) and Solved (1) instances on the calibration set. A distinct shift in the Mean values across all metrics (LGC, MC, TC, BWC) indicates a clear decision boundary, supporting the hypothesis that confidence is a strong predictor of correctness (on a stratified subset of 50 samples).}
\label{tab:confidence_stats}
\end{table*}

\subsection{Details of Confidence Metrics}
\label{sec:conf_metrics}
To construct the feature vector $C(x_i)$, we adopt an uncertainty quantification approach based on \citet{fu2025deepthinkconfidence}. Rather than relying on raw probability margins, we utilize negative token entropy as the fundamental unit of confidence, defined as \textit{Token Confidence} metric $c_t$, defined as the negative average log-probability of the top-$k$ at position $i$:
\begin{equation}
    c_i = - \frac{1}{k} \sum_{j=1}^{k} \log P_i(j)
\end{equation}
where $P_i(j)$ is the probability of the $i$-th token. Under this formulation, values close to 0 indicate model uncertainty (peaked distributions), while larger values indicate high certainty.

\paragraph{Lowest Group Confidence (LGC)}
LGC detects localized reasoning collapses by identifying the "weakest link" in the generation. We compute the average confidence over sliding windows of size $W$ and select the minimum:
\begin{equation}
    \text{LGC} = \min_{i} \left( \frac{1}{W} \sum_{j=i}^{i+W-1} c_j \right)
\end{equation}

\paragraph{Mean Confidence (MC)}
MC serves as a proxy for global model self-assurance, averaging entropy across the full sequence length $T$:
\begin{equation}
    \text{MC} = \frac{1}{T} \sum_{t=1}^{T} c_t
\end{equation}

\paragraph{Tail Confidence (TC)}
Acknowledging that reasoning errors often compound (or resolve) towards the end of a chain, TC isolates the confidence of the final $W$ tokens:
\begin{equation}
    \text{TC} = \frac{1}{W} \sum_{t=T-W+1}^{T} c_t
\end{equation}

\paragraph{Bottom-K\% Group Confidence (BWC)}
BWC aggregates the lowest $K\%$ of window scores to distinguish between isolated token noise and systemic hallucination. Let $S_K$ be the subset of window scores $\{G_i\}$ comprising the bottom $K$-th percentile:
\begin{equation}
    \text{BWC} = \frac{1}{|S_K|} \sum_{G \in S_K} G
\end{equation}

\subsection{Statistical Analysis of Confidence Metrics}
\label{sec:stat_analysis}

To validate the fundamental hypothesis that intrinsic model uncertainty correlates with solution correctness, we performed a statistical analysis of the confidence metrics generated during the warm-up phase. \autoref{tab:confidence_stats} details the distributional properties (Mean, Standard Deviation, and Skewness) of the four proposed metrics: Lowest Group Confidence (LGC), Mean Confidence (MC), Tail Confidence (TC), and Bottom-Weighted Confidence (BWC) - segregated by ground truth outcomes.

The data reveals a significant distributional shift between Unsolved (0) and Solved (1) instances. Most notably, the \textbf{Mean} values for successful generations are consistently higher across all metrics, with gaps ranging from approximately 1.7 to 2.4 units of negative entropy. For instance, LGC shifts from a mean of 5.86 in unsolved cases to 7.98 in solved instances. Additionally, the \textbf{Skewness} is markedly higher in unsolved instances for metrics like LGC (2.04 vs. 0.93) and BWC (1.71 vs. 0.72), suggesting that heavy tails of extreme uncertainty characterize incorrect generations.

This statistical separation signifies that the "confidence signature" of the small model is not random noise but a reliable predictor. The distinct margins between the class means provide the necessary signal for our Decision Tree router to establish effective decision boundaries, thereby justifying the use of these specific metrics for adaptive model escalation.
\subsection{Justification for Non-Linear Routing (Pilot Study)}

\begin{table}[h]
\centering
\resizebox{\columnwidth}{!}{%
\begin{tabular}{@{}lccc@{}}
\toprule
\textbf{Setting} & \textbf{Cost} & \textbf{Accuracy} & \textbf{Eff.} \\ \midrule
4B Iterative (Lower Bound) & 0.46 & 62.15 & 135.1 \\
\textbf{32B Iterative (Upper Bound)} & \textbf{3.07} & \textbf{76.20} & \textbf{24.8} \\ \midrule
Threshold Switch (z-score) & 2.08 & 70.07 & 33.7 \\
Random Sampling (35:65) & 2.03 & 72.23 & 35.6 \\
\textbf{Decision Tree (Ours)} & \textbf{2.02} & \textbf{74.12} & \textbf{36.7} \\ \bottomrule
\end{tabular}%
}
\caption{\textbf{Pilot Study on Router Selection ($N=50$).} A comparison of routing mechanisms on a stratified subset of LiveCodeBench. The \textbf{Decision Tree} achieves the highest efficiency among the mixed-model strategies, validating the effectiveness of non-linear confidence routing.}
\label{tab:pilot_ablation}
\end{table}

To validate our architectural choice of a Decision Tree over simpler heuristics, we conducted a preliminary ablation study on a stratified subset of the dataset ($N=50$). We compared our proposed Decision Tree router against a \textit{Threshold-based Switch}, which aggregates confidence metrics into a scalar z-score and applies a linear cutoff.

As detailed in Table \ref{tab:pilot_ablation}, the linear Threshold baseline achieved an accuracy of 0.70 with a compute cost of 2.08 compute cost. Conversely, the Decision Tree router achieved a superior accuracy of 0.74 while maintaining a lower computational cost (2.02 compute cost). This performance gap indicates that the "confidence signature" of hallucinations is non-linear; the Decision Tree successfully captures all-to-one interactions between metrics (e.g., high mean confidence but low tail confidence) that a simple linear threshold fails to distinguish. Consequently, the Decision Tree was selected as the routing mechanism for the full evolutionary framework.


\subsection{Hyperparameters}
\label{sec:hyperparameters}
\begin{table}[h]
\centering
\resizebox{\columnwidth}{!}{%
\begin{tabular}{@{}llc@{}}
\toprule
\textbf{Category} & \textbf{Parameter} & \textbf{Value} \\ \midrule
\textbf{General} & Max Iterations & 8 \\
& Checkpoint Interval & 1 \\
& Random Seed & 42 \\ \midrule
\textbf{LLM Ensemble} & Temperature & 0.6 \\
& Top-p & 0.95 \\
& Max Tokens & 20000 \\
& Timeout & 3600s \\
& Model & Qwen3-32B, Qwen3-4B \\ \midrule
\textbf{Database / Evolution} & Population Size & 8 \\
& Archive Size & 3 \\
& Number of Islands & 2 \\
& Elite Selection Ratio & 0.3 \\
& Exploration Ratio & 0.2 \\
& Exploitation Ratio & 0.5 \\
& Feature Dimensions & Pass Rate, Complexity \\
& Feature Bins & 5 \\
& Migration Interval & 10 \\
& Migration Rate & 0.15 \\ \midrule
\textbf{Evaluator} & Evaluation Timeout & 1200s \\
& Parallel Evaluations & 32 \\
& Max Retries & 4 \\
& Cascade Evaluation & False \\ \midrule
\textbf{Confidence Calculation Window} & Tail Confidence & 2048 \\
& Bottom Grouped Confidence Window & 2048 \\
& Tail Grouped Confidence Window & 2048 \\ \bottomrule
\end{tabular}%
}
\caption{\textbf{Evolutionary Framework Configuration.} Hyperparameters used for the updated evolutionary search and inference environment.}
\label{tab:config_params_updated}
\end{table}

\begin{figure*}[t]
    \centering
    \includegraphics[width=1\linewidth]{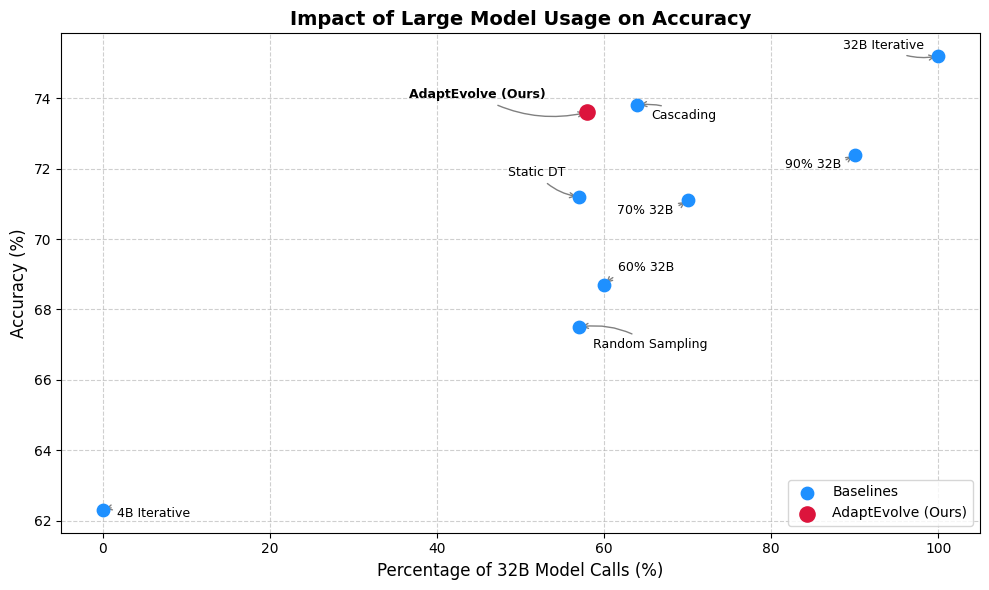}
    \caption{\textbf{Impact of Large Model Usage on Accuracy (LiveCodeBench).} We plot the percentage of calls routed to the 32B model against the final accuracy. \textit{AdaptEvolve} (Red) achieves a significantly better trade-off than the Cascading baseline and Random Sampling, attaining near-peak accuracy with only 58\% large-model usage.}
    \label{fig:accuracy_vs_32b}
\end{figure*}

\begin{table}[h!]
\centering

\resizebox{\columnwidth}{!}{%
\begin{tabular}{@{}lllll@{}}
\toprule
\textbf{Current Model} & \textbf{Current Pass Rate ($P$)} & \textbf{Ground Truth Other Model} & \textbf{Target Label} & \textbf{Reasoning} \\ \midrule
4B & $P \ge 1.0$ & (Irrelevant) & \textbf{0} (Use 4B) & 4B solved efficiently; maintain status quo. \\
4B & $P < 1.0$ & 32B Solvable & \textbf{1} (Use 32B) & 4B failed, but 32B is known to succeed. \\
4B & $P < 1.0$ & 32B Not Solvable & \textbf{0} (Use 4B) & Hard problem; larger model implies waste. \\ \midrule
32B & $P \ge 1.0$ & 4B Solvable & \textbf{0} (Use 4B) & Both solve it; switch down to save cost. \\
32B & $P \ge 1.0$ & 4B Not Solvable & \textbf{1} (Use 32B) & Only 32B solves it; maintain larger model. \\
32B & $P < 1.0$ & (Irrelevant) & \textbf{0} (Use 4B) & 32B failed; revert to base model. \\ \bottomrule
\end{tabular}%
}
\caption{Labeling Logic for Adaptive Model Switching ($P = 1.0$)}
\label{tab:router_logic}
\end{table}

\end{document}